\pdfoutput=1

\documentclass[11pt]{article}

\usepackage[final]{acl}

\usepackage{times}
\usepackage{latexsym}

\usepackage[T1]{fontenc}

\usepackage[utf8]{inputenc}

\usepackage{microtype}

\usepackage{inconsolata}

\usepackage{graphicx}

\usepackage{booktabs}
\usepackage{amsmath}
\usepackage{gb4e}
\noautomath  
\usepackage{tipa}

\title{Hybrid Neural-LLM Pipeline for Morphological Glossing in Endangered Language Documentation: A Case Study of Jungar Tuvan}

\author{
  Siyu Liang$^{1}$,
  Talant Mawkanuli$^{2}$, 
  Gina-Anne Levow$^{1}$\\
  $^{1}$ Department of Linguistics, University of Washington \\
  $^{2}$ Department of Middle Eastern Languages and Cultures, University of Washington \\
  \texttt{\{liangsy, tmawkan, levow\}@uw.edu}
}

\begin{document}
\maketitle

\begin{abstract}
Interlinear glossed text (IGT) creation remains a major bottleneck in linguistic documentation and fieldwork, particularly for low-resource morphologically rich languages. We present a hybrid automatic glossing pipeline that combines neural sequence labeling with large language model (LLM) post-correction, evaluated on Jungar Tuvan, a low-resource Turkic language. Through systematic ablation studies, we show that retrieval-augmented prompting provides substantial gains over random example selection. We further find that morpheme dictionaries paradoxically hurt performance compared to providing no dictionary at all in most cases, and that performance scales approximately logarithmically with the number of few-shot examples. Most significantly, our two-stage pipeline combining a BiLSTM-CRF model with LLM post-correction yields substantial gains for most models, achieving meaningful reductions in annotation workload. Drawing on these findings, we establish concrete design principles for integrating structured prediction models with LLM reasoning in morphologically complex fieldwork contexts. These principles demonstrate that hybrid architectures offer a promising direction for computationally light solutions to automatic linguistic annotation in endangered language documentation. 
\end{abstract}

\section{Introduction}
Interlinear glossed text (IGT) is essential for linguistic documentation and preservation, aligning language transcriptions with morpheme segmentation, glosses, and translations \citep{lehmann_data_2004}. Despite its importance, IGT creation remains highly labor-intensive, creating bottlenecks in language documentation projects \citep{chelliah_handbook_2010}. Recent advances in neural sequence models and large language models offer new possibilities for automated IGT generation, yet each approach has limitations: structured models lack flexibility and world knowledge, while LLMs struggle with consistency and require extensive in-context examples or expensive fine-tuning, all inaccessible in most real-life use cases. 

We present a hybrid pipeline that combines a BiLSTM-CRF model for initial gloss prediction with LLM-based post-correction, evaluated on Jungar Tuvan (henceforth Tuvan), a morphologically complex Turkic language. Through systematic experiments across four LLMs and multiple design choices, we demonstrate that this two-stage approach substantially improves over the BiLSTM baseline for most models. Our ablation studies reveal key design principles: retrieval-augmented prompting significantly outperforms random example selection; morpheme dictionaries generally hurt performance for most models; and optimal few-shot parameters range from five to fifteen examples. 

Our contributions are the following: (1) we present a hybrid architecture combining structured prediction with LLM reasoning for automatic glossing; (2) we carry out comprehensive ablation studies establishing design principles for retrieval strategies, glossary configurations, and few-shot scaling; (3) we provide comparative evaluation of model performance across generation versus correction tasks, providing evidence-based guidance in fieldwork contexts. 

\section{Related Work}

\subsection{IGT and Language Documentation}
IGT serves as a standard format in field linguistics, encoding source language transcriptions, morphological segmentation, gloss labels, and free translations \citep{booij_interlinear_2004, comrie_leipzig_2008}. For many endangered and low-resource languages, IGT represents the primary digitized documentation \citep{chelliah_handbook_2010, hargus_glossing_2020}. 

Recent work has developed tools for IGT extraction from grammatical descriptions \citep{schenner_extracting_2016, round_automated_2020, nordhoff_imtvault_2022} and multi-modal IGT generation from speech \citep{he_wav2gloss_2024}. The SIGMORPHON 2023 shared task on automatic IGT generation \citep{ginn_findings_2023} further stimulated interest in computational approaches to glossing, leading to subsequent work on large-scale IGT modeling and evaluation, including pretrained language models for glossing \citep{ginn_glosslm_2024} and LLM-based prompting approaches for low-resource IGT generation \citep{elsner_prompt_2025}. While these efforts demonstrate steady progress on benchmark datasets, they have not yet displaced existing fieldwork practices; in most documentation projects, IGT creation remains largely manual, relying on tools such as ELAN and FLEx \citep{wittenburg_elan_2006, sil_international_fieldworks_2025}. Traditional workflows also include rule-based morphological parsers and deterministic dictionary lookup within tools like FLEx, which supports semi-automatic glossing and lexicon building from texts; our approach is intended to complement rather than replace such methods.

\subsection{Automatic Morphological Analysis and Glossing}
Traditional approaches to automatic glossing employ structured prediction models. Sequence-to-sequence architectures have been applied to morphological segmentation  \citep{ruzsics_neural_2017, liu_morphological_2021, rice_tams_2024}, while CRF-based models have proven effective for morphological tagging \citep{buys_cross-lingual_2016, malaviya_neural_2018}. BiLSTM-CRF architectures in particular balance local pattern recognition with global constraints \citep{ma_end--end_2016, cotterell_cross-lingual_2017}, achieving strong performance on sequence labeling tasks in morphologically rich languages.

Recent work specifically targeting IGT generation has explored neural encoder-decoder models with translation data \citep{zhao_automatic_2020}, CRF-based approaches for low-resource scenarios \citep{barriga_martinez_automatic_2021, okabe_towards_2023}, and lightweight models using structured linguistic representations \citep{shandilya_lightweight_2023}. \citet{moeller_igt2p_2020} demonstrate how IGT can support downstream morphological analysis tasks. However, these models require substantial annotated corpora and struggle with rare morphemes and novel combinations. The recent work by \citet{rice_interdisciplinary_2025} also identifies significant gaps between computational morphology research outputs and real-world language documentation needs, highlighting the importance of user-centered design. 

\subsection{LLMs for Linguistic Annotation}
Large language models have shown promise for linguistic annotation tasks in low-resource settings. Recent work \citep{ginn_can_2024, elsner_prompt_2025} explore LLM-based gloss prediction and prompting strategies for IGT, demonstrating that prompt design and example selection substantially affect performance. \citet{zhang_hire_2024} show that providing dictionaries and grammar sketches enables translation for unseen languages. \citet{yang_linggym_2025} evaluate models on metalinguistic reasoning using reference grammars and IGT, though their benchmark relies on curated reference grammar data without the full fieldwork contexts. LLM post-correction and refinement steps are also widely explored across NLP tasks as cascaded or post-editing stages \citep{zouhar_neural_2021, izacard_atlas_2023}.

Among recent studies, few-shot prompting has emerged as a key technique for adapting LLMs to specialized tasks. Studies demonstrate that careful selection and presentation of in-context examples significantly impacts performance \citep{logan_iv_cutting_2022, winata_language_2021}, with retrieval-based example selection often outperforming random selection \citep{stahl_exploring_2024}. However, LLMs face challenges in low-resource settings: they require extensive in-context examples (increasing inference cost), struggle with paradigmatic consistency, and lack the inductive biases of structured sequence models.

\subsection{Hybrid and Multi-Stage Architectures}
Hybrid approaches combining multiple model types have proven effective across NLP tasks. Retrieval-augmented generation (RAG) enhances LLMs by dynamically incorporating relevant examples \citep{lewis_retrieval-augmented_2021, jiang_active_2023}, with recent work exploring specialized RAG architectures for domain adaptation \citep{siriwardhana_improving_2023, yu_retrieval-augmented_2022}. Cascaded architectures leverage specialized models for different subtasks \citep{izacard_atlas_2023}, while post-processing steps that refine outputs using external knowledge have shown consistent gains \citep{zouhar_neural_2021}. Our work extends these ideas to morphological annotation, proposing a two-stage pipeline where a BiLSTM-CRF provides initial structure and an LLM refines predictions through contextual inference and consistency checks.

\section{Data}

\subsection{Language and Corpus}
Tuvan is a Turkic language spoken in the Republic of Tuva of the Russian Federation, Mongolia, and the Xinjiang Uyghur Autonomous Region of China, with approximately 280,000 speakers across these regions \citep{harrison_grammar_2002}. The present study focuses on the variety of Jungar Tuvan, spoken in the Altay region of Xinjiang, China. We treat Jungar Tuvan as a low-resource variety used in documentation contexts; we do not adjudicate its formal endangerment status in this paper. Jungar Tuvan shares the core typological properties of Tuvan—canonical agglutinative morphology, extensive case marking (nominative, accusative, genitive, dative, locative, ablative, comitative), complex aspectual systems, and productive derivational morphology—while also exhibiting vowel harmony and consonant alternations that create allomorphic variation \citep{mawkanuli_phonology_1999, mawkanuli_jungar_2005}.

Our corpus comprises 895 IGT-annotated sentences drawn from data collected during fieldwork in Xinjiang, China from the 1987 to 1995. The data span 40 recording sessions across conversational registers and narratives. All IGT annotations were produced manually following a consistent project-specific schema informed by typological conventions \citep{booij_interlinear_2004, comrie_leipzig_2008}. Glosses distinguish lexical items (e.g., \textit{money}, \textit{give}) from grammatical morphemes (e.g., \textsc{dat}, \textsc{1sg}, \textsc{prs}). 

Table~\ref{tab:corpus_stats} summarizes corpus statistics. The tagset comprises 240 unique gramamtical morpheme labels (e.g., \textsc{1sg}, \textsc{pst}), and 1258 unique content word glosses. 

\begin{table}[h]
\centering
\small
\begin{tabular}{lr}
\toprule
\textbf{Metric} & \textbf{\#} \\
\midrule
Total sentences & 895 \\
Narratives & 40 \\
Unique grammatical morphemes & 240 \\
Unique content morphemes & 1258 \\
Average words per sentence & 8.38 (± 5.71) \\
Average morphemes per word & 1.69 (± 0.30) \\
\bottomrule
\end{tabular}
\caption{Corpus statistics for the Tuvan fieldwork dataset.}
\label{tab:corpus_stats}
\end{table}

Example~\ref{ex:tuvan} illustrates a representative IGT instance from the corpus, demonstrating the alignment structure our models must learn.

\begin{exe}
	\ex\label{ex:tuvan}
	\gll jïlgï-nan \quad iyi \quad joon \quad bar \\
			 horse-\textsc{abl} \quad two \quad big \quad \textsc{exist} \\
	\glt ``(We) have two big horses.''
\end{exe}

\subsection{Data Split}
We perform a train-test split at the document level, allocating approximately 85\% of sentences (760) to training and 15\% (135) to testing. To avoid information leakage, no segments of the same narrative appear in both splits, ensuring that models cannot exploit discourse-level or speaker-specific patterns from related utterances. We further verify the absence of near-duplicate sentences using character-level TF-IDF (term frequency–inverse document frequency) cosine similarity with a threshold of 0.95.

\section{Methodology}

\subsection{Task Formalization}
We frame glossing as a structured prediction problem: given a hyphen-segmented Tuvan utterance $\mathbf{x} = (x_1, x_2, \ldots, x_n)$ where each $x_i$ represents a morpheme, produce a parallel sequence of gloss labels $\mathbf{y} = (y_1, y_2, \ldots, y_n)$ where each $y_i$ is drawn from a tagset. We assume gold morpheme boundaries and do not use translations in the glossing model; segmentation and glossing are treated as separate steps. We evaluate using token-level accuracy, defined as:
\begin{equation}
\text{Accuracy} = \frac{1}{N}\sum_{i=1}^{N} [\hat{y}_i = y_i]
\end{equation}
where $N$ is the total number of morphemes in the test set, $\hat{y}_i$ is the predicted gloss, and $y_i$ is the reference gloss. This metric directly reflects annotation workload reduction: higher accuracy means fewer manual corrections required.

\subsection{BiLSTM-CRF Model}
Our baseline employs a two-layer bidirectional LSTM with CRF decoding, widely used for sequence labeling \citep{lample_neural_2016, ma_end--end_2016, huang_bidirectional_2015}. The model uses 100-dimensional character-level embeddings (randomly initialized and trained from scratch), 128-dimensional hidden layers, and learns to predict gloss labels for segmented morphemes. We train for up to 100 epochs with early stopping (patience=10) on validation loss. This architecture captures local morphological patterns but cannot leverage broader linguistic knowledge or rare paradigms not well-represented in training data. 

\subsection{LLM Configuration and Prompting}
We evaluate four LLMs: \texttt{deepseek-v3.2-exp} \citep{deepseek-ai_deepseek-r1_2025}, \texttt{qwen3-max} \citep{yang_qwen3_2025}, \texttt{gpt-4o-mini} \citep{openai_gpt-4_2024}, and \texttt{gemma-3-27b-it} \citep{kamath_gemma_2025}. All models use greedy decoding with temperature zero for deterministic outputs. Prompts follow a consistent template: natural language instruction, $k$ retrieved examples showing morpheme segmentation and gloss pairs, optional morpheme dictionary, and the test input. For the hybrid pipeline, prompts additionally include the BiLSTM prediction as a hypothesis to correct, framed as a ``rough initial attempt'' requiring verification. Complete prompt templates for all experiments are provided in Appendix~\ref{sec:appendix}. Figure~\ref{fig:methodology} illustrates our hybrid pipeline architecture.

\begin{figure}[t]
\centering
\includegraphics[width=\columnwidth]{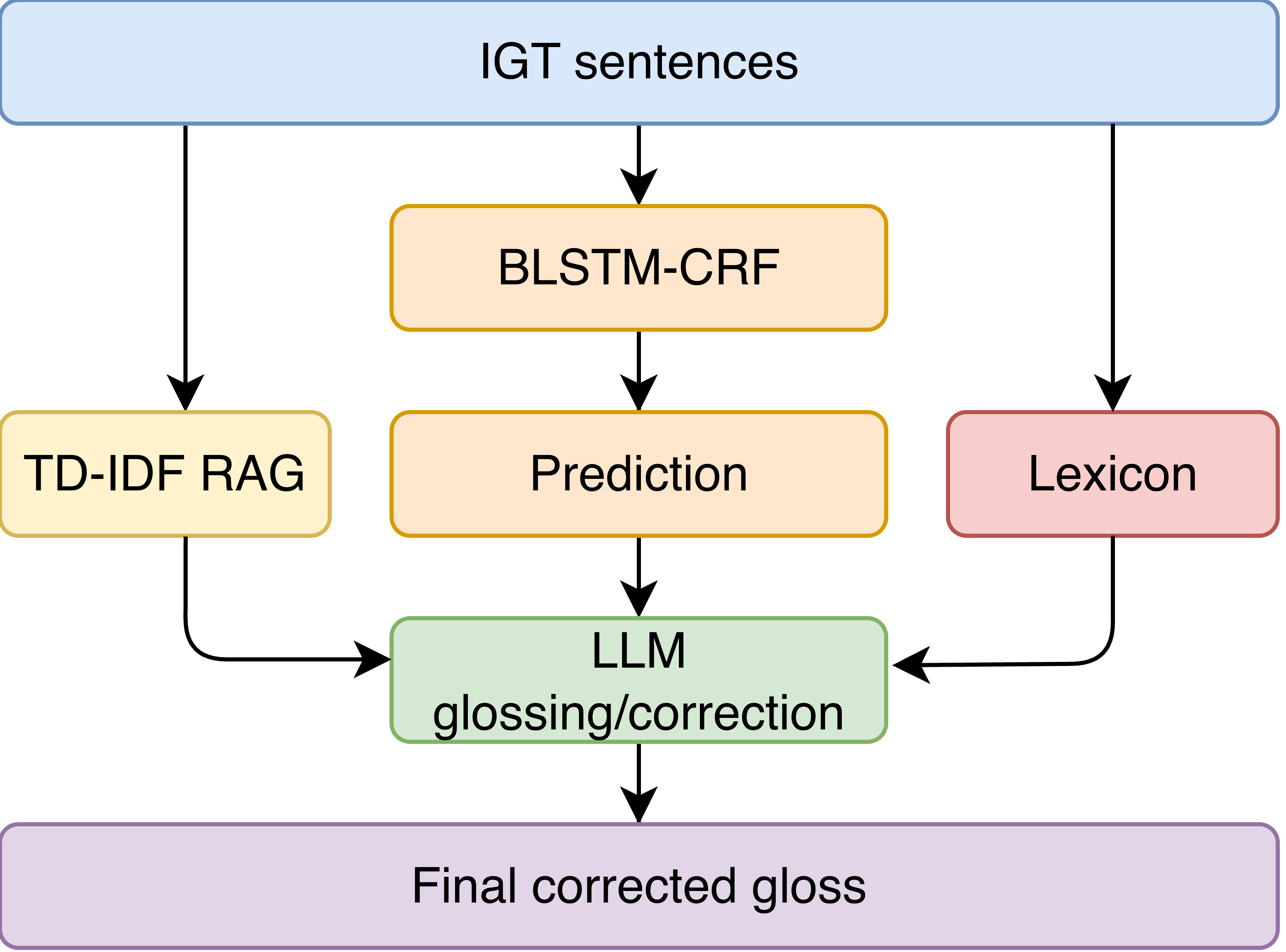}
\caption{Hybrid pipeline combining BiLSTM-CRF structured prediction with LLM post-correction using retrieval-augmented prompting.}
\label{fig:methodology}
\end{figure}

\subsection{Experimental Design}
We conduct several experiments testing both LLM generation and hybrid correction. In Experiment 1 (Retrieval vs.\ Random Selection), we compare character-level TF-IDF (term frequency–inverse document frequency) cosine similarity-based retrieval against uniform random sampling for selecting three in-context examples to explore the effect of similarity-based retrieval. Retrieval operates at the sentence level: for each test sentence, we retrieve the most similar training sentences based on their Tuvan source text representations. Experiment 2 (N-Shot Scaling) varies example count from 1 to 20 with RAG and no glossary, mapping the accuracy-cost tradeoff for RAG LLM generation.

Experiment 3 (Glossary Ablation) tests four glossary configurations---none, top-100 most frequent morphemes, all grammatical morphemes, and the entire 1,498-pair dictionary---all using three-shot RAG to reveal whether partial dictionaries help or hinder performance. The glossary is provided to the LLM within the prompt as a plain-text key:value list (Appendix~\ref{sec:appendix}), rather than as a deterministic lookup table. Finally, Experiment 4 (Hybrid Pipeline) evaluates BiLSTM plus LLM correction with varying n-shot counts including a zero-shot condition that tests whether LLMs can correct predictions without in-context examples. These ablations correspond to fieldwork-relevant choices about retrieval quality, example budget, and availability of lexical resources. Detailed prompt templates for RAG LLM generation (Experiments 1--3) and hybrid correction (Experiment 4) are provided in Appendix~\ref{sec:appendix}. 

\section{Results}

\subsection{Baseline: BiLSTM-CRF Performance}
Our BiLSTM-CRF baseline achieves 0.474 token accuracy on the test set with training data of 760 sentences. The model learns frequent morphological patterns (case markers, possessives, tense/aspect) but struggles with infrequent lexical morphemes and combinations of grammatical morphemes not attested in the training data. Error analysis reveals that 0.38 of errors involve lexical items appearing fewer than 5 times in training, and 0.24 involve grammatical morphemes in novel combinations.

\subsection{Experiment 1: Retrieval vs.\ Random Selection}
Table~\ref{tab:retrieval_all_models} shows the effect of retrieval enhancement across all four LLMs using 3-shot prompting without glossary.

\begin{table}[h]
\centering
\small
\begin{tabular}{lcc}
\toprule
\textbf{Model} & \textbf{Random} & \textbf{RAG} \\
\midrule
\texttt{deepseek-v3.2-exp} & 0.118 & \textbf{0.506} \\
\texttt{qwen3-max}     & 0.062  & 0.381 \\
\texttt{gpt-4o-mini}   & 0.103 & 0.396 \\
\texttt{gemma-3-27b-it}   & 0.068 & 0.344 \\
\bottomrule
\end{tabular}
\caption{Retrieval-augmented prompting (RAG) vs.\ random example selection across four LLMs (3-shot, no glossary). All models show meaningful improvement with RAG.}
\label{tab:retrieval_all_models}
\end{table}


Retrieval-augmented generation provides meaningful improvements across all models: +0.388 for \texttt{deepseek-v3.2-exp}, +0.319 for \texttt{qwen3-max}, +0.293 for \texttt{gpt-4o-mini}, and +0.276 for \texttt{gemma-3-27b-it}. \texttt{deepseek-v3.2-exp} achieves the highest absolute accuracy with RAG (0.506), while all models show substantial gains from retrieval. The consistent gains across architectures demonstrate that similarity-based example selection is beneficial for morphological glossing tasks.

\subsection{Experiment 2: N-Shot Scaling}
Using the same TF-IDF-based retrieval from Experiment 1, we vary the number of retrieved examples from 1 to 20 without providing any glossary. Figure~\ref{fig:exp2_nshot} shows performance scaling with example count (see Table~\ref{tab:nshot_scaling} in Appendix~\ref{sec:appendix-tables} for detailed values).

\begin{figure}[t]
\centering
\includegraphics[width=\columnwidth]{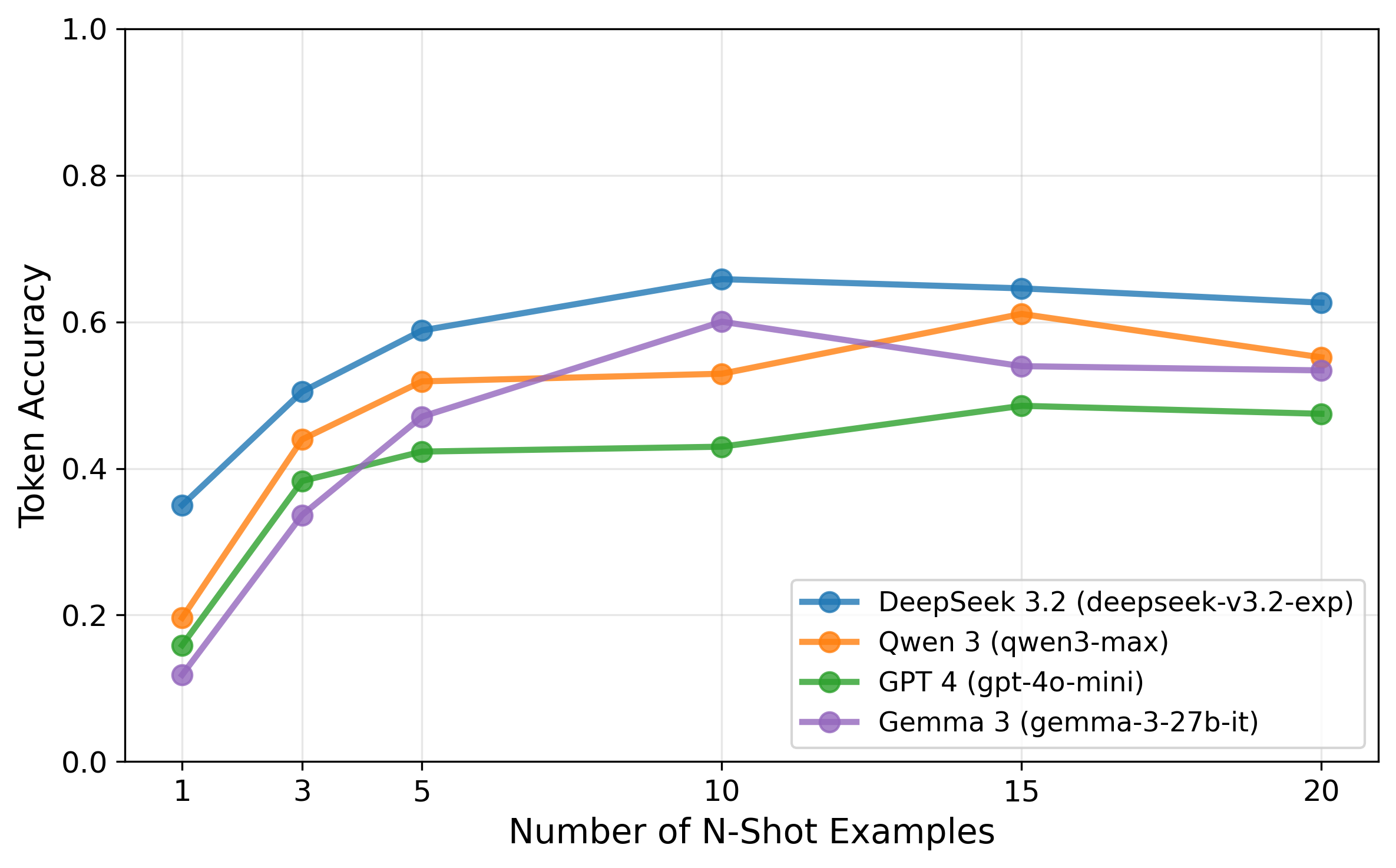}
\caption{Experiment 2: n-shot scaling curves for RAG LLM generation. Performance scales approximately logarithmically with example count, plateauing around n=10--15 for most models. The BiLSTM baseline (0.474) is provided in the text for reference.}
\label{fig:exp2_nshot}
\end{figure}

Performance scales approximately logarithmically with example count. \texttt{deepseek-v3.2-exp} peaks at n=10 (0.658), then slightly declines at n=15 (0.646) and n=20 (0.626), while \texttt{qwen3-max} shows continued gains up to n=15 (0.611). \texttt{gpt-4o-mini} peaks at n=15 (0.486), and \texttt{gemma-3-27b-it} achieves 0.600 at n=10 before declining to 0.534 at n=20. The consistent pattern across models indicates diminishing marginal returns beyond 10 to 15 examples, with some models showing degradation at higher values. This decline may reflect either model saturation (distraction from excessive context) or retrieval quality degradation (less similar examples as the pool expands). We use character-level TF-IDF cosine similarity for retrieval; alternative similarity measures such as edit distance or contextualized embeddings might exhibit different scaling behavior, though we leave this investigation to future work. These results suggest practical operating points around n=5 to 10 for cost-sensitive applications and n=10 to 15 for maximum accuracy.

\subsection{Experiment 3: Glossary Ablation}
To assess the impact of morpheme dictionaries on performance, we test four glossary configurations using 3-shot RAG: None (no dictionary provided), Top-100 (the 100 most frequent morpheme-gloss pairs), Grammatical (all 240 grammatical morpheme-gloss pairs), and Entire (the complete 1,498-pair dictionary including both grammatical and lexical morphemes). Figure~\ref{fig:exp3_glossary} shows the results (see Table~\ref{tab:glossary_all_models} in Appendix~\ref{sec:appendix-tables} for detailed values).

\begin{figure}[t]
\centering
\includegraphics[width=\columnwidth]{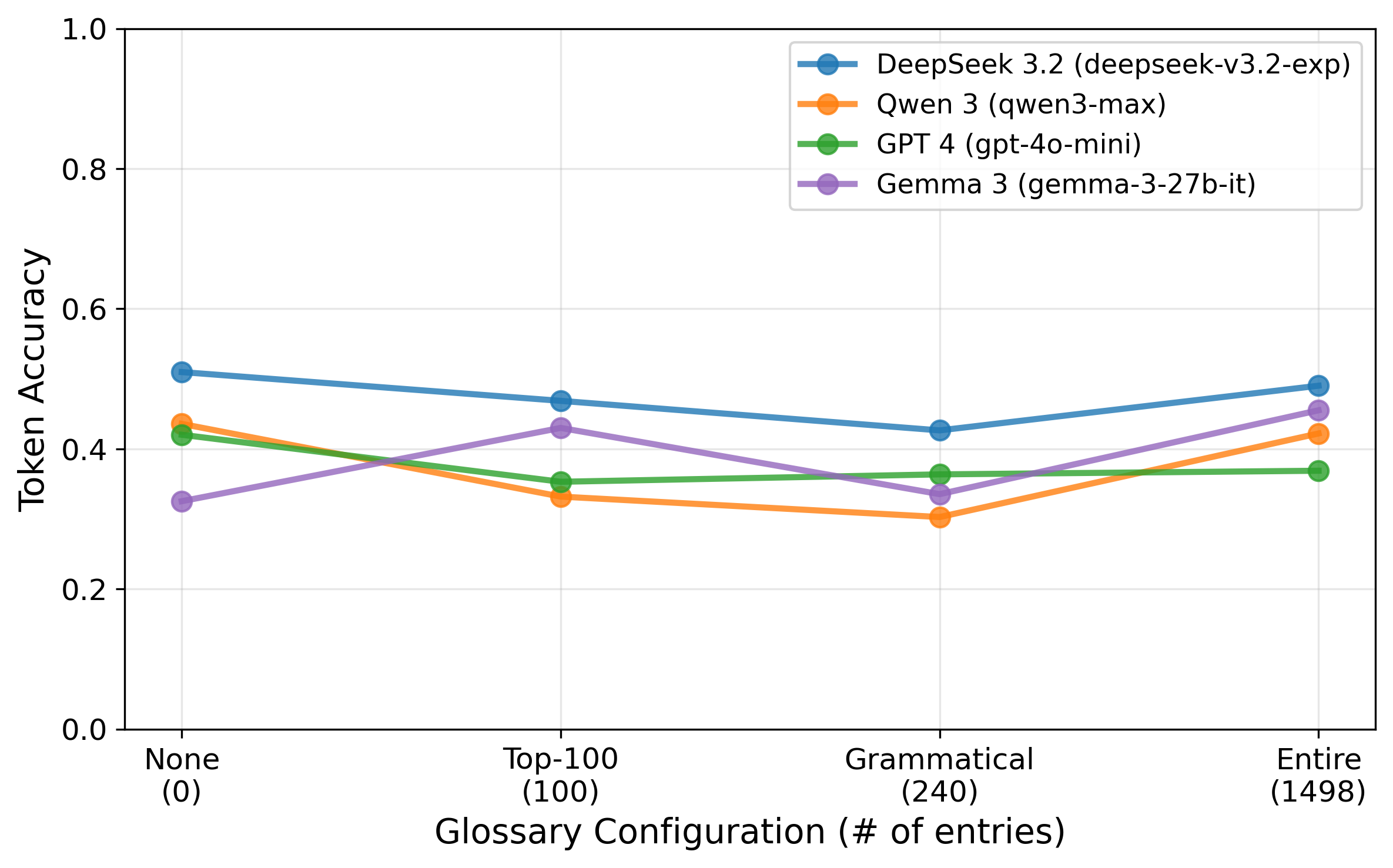}
\caption{Experiment 3: glossary ablation results. Partial glossaries (Top-100, Grammatical) hurt performance compared to no glossary, while complete glossaries show modest gains. The negative effect suggests models are usually distracted by morphological information. The BiLSTM baseline (0.474) is provided in the text for reference.}
\label{fig:exp3_glossary}
\end{figure}

Counter-intuitively, providing morpheme dictionaries generally hurts performance. Partial glossaries consistently degrade accuracy across all models: Top-100 causes drops ranging from 0.041 to 0.104, while Grammatical shows similar or worse declines. Even the complete 1,498-pair dictionary fails to help for most models, with \texttt{deepseek-v3.2-exp}, \texttt{qwen3-max}, and \texttt{gpt-4o-mini} all performing worse with the entire glossary than with none (losses of 0.019, 0.014, and 0.051 respectively). Only \texttt{gemma-3-27b-it} benefits from dictionary information, achieving 0.455 with the entire glossary (+0.130 over None). 

We lack direct evidence about whether models actually consult dictionary entries versus simply becoming distracted by additional prompt material. The degradation could reflect either inappropriate reliance on dictionary lookups for context-dependent glossing decisions, or models' difficulty balancing multiple information sources (retrieved examples, dictionaries, and morphological patterns). The no-glossary condition forces models to extract morphological structure from aligned examples, which appears more effective than dictionary consultation for most architectures, though we cannot definitively isolate the causal mechanism without fine-grained analysis of model predictions.

\subsection{Experiment 4: Hybrid Pipeline Performance}
\begin{figure}[t]
\centering
\includegraphics[width=\columnwidth]{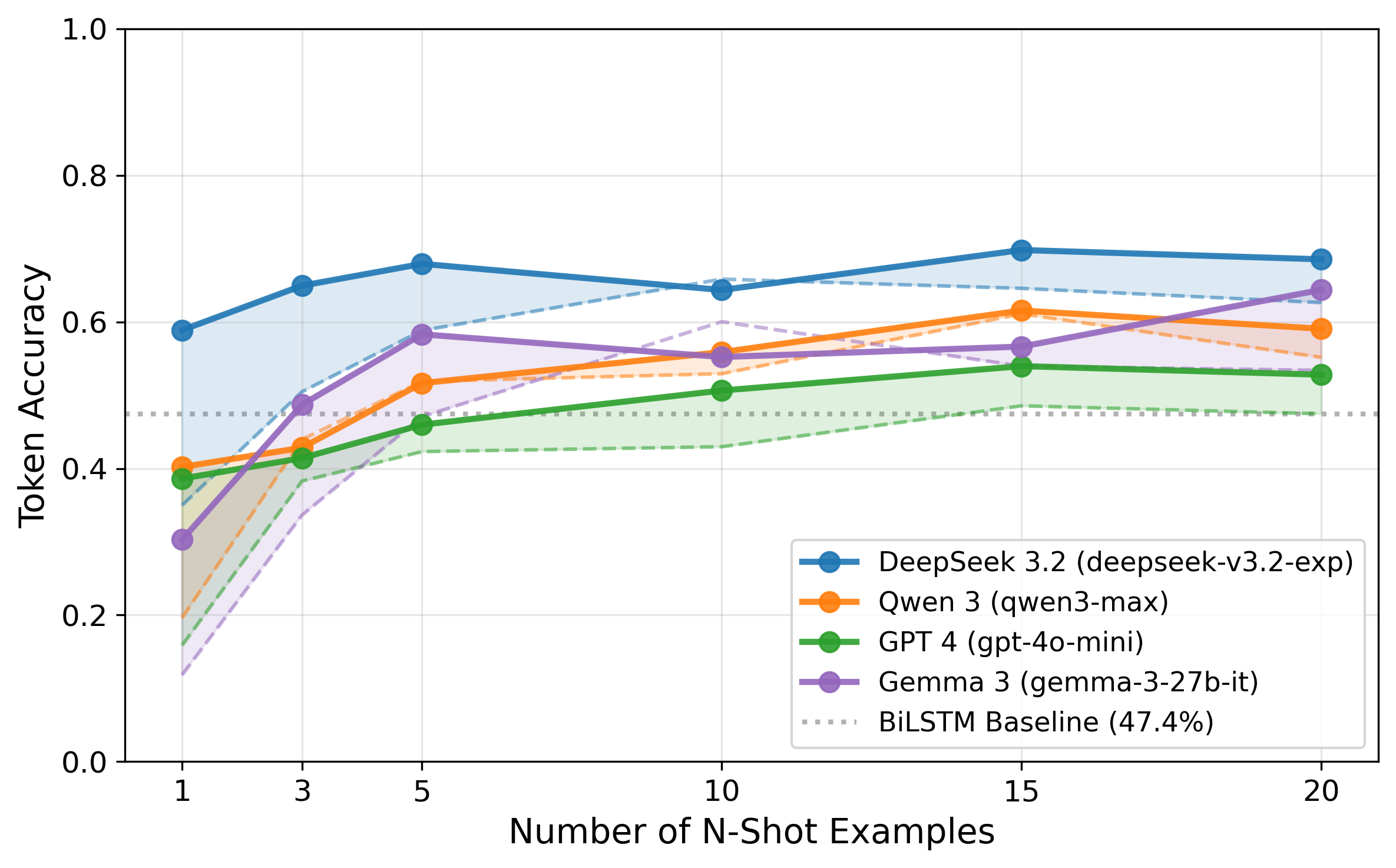}
\caption{Experiment 4: hybrid pipeline improvement over RAG LLM generation. Solid lines show hybrid accuracy (BiLSTM + LLM correction), dashed lines show pure N-Shot baseline from Experiment 2, and shaded areas indicate improvement. The hybrid approach consistently improves performance across all four models, particularly in low-shot scenarios (n=1--5).}
\label{fig:exp4_improvement}
\end{figure}

We then evaluate the hybrid pipeline using the same TF-IDF retrieval as in Experiments 1 and 2, but providing the BiLSTM-CRF predictions as initial hypotheses for LLM correction. We test with n=1, 3, 5, 10, 15, 20 retrieved examples, all without glossaries. Figure~\ref{fig:exp4_improvement} shows improvement from the hybrid pipeline over RAG LLM generation across varying n-shot counts for all four models.

The hybrid pipeline substantially outperforms pure BiLSTM predictions across all models. \texttt{gemma-3-27b-it} achieves 0.644 at n=20, a +0.170 gain over the BiLSTM baseline (0.474) and +0.110 improvement over its pure generation performance (0.534 at n=20). \texttt{deepseek-v3.2-exp} reaches 0.698 at n=15 (+0.224 over BiLSTM, +0.052 over pure generation), \texttt{qwen3-max} achieves 0.616 (+0.142 over BiLSTM, +0.005 over pure generation), and \texttt{gpt-4o-mini} reaches 0.540 (+0.066 over BiLSTM, +0.054 over pure generation), demonstrating that the hybrid approach benefits all models across performance tiers. 

The improvements are particularly substantial in low-shot scenarios: at n=1, \texttt{gemma-3-27b-it} shows +0.185 improvement over pure generation (0.118 → 0.303), \texttt{deepseek-v3.2-exp} improves by +0.058, and \texttt{gpt-4o-mini} by +0.063. This demonstrates that the BiLSTM provides valuable structural guidance when few examples are available. As n increases, the gains diminish but remain consistent across all models, with \texttt{gemma-3-27b-it} showing +0.110 improvement even at n=20.

\subsection{Error Analysis}
To understand where improvements originate, we analyze errors by morpheme type. Comparing BiLSTM baseline against the best-performing hybrid configuration (\texttt{deepseek-v3.2-exp} with $n=10$ RAG examples), we find asymmetric improvements.

The BiLSTM baseline achieves 0.923 accuracy on 168 grammatical morphemes but only 0.479 on 213 lexical morphemes. This confirms that structured models learn morphological paradigms effectively but struggle with lexical gaps. The hybrid pipeline improves lexical accuracy to 0.682 (+0.204 absolute gain) while grammatical accuracy drops slightly to 0.866 ($-$0.057). The substantial lexical gains outweigh minor grammatical losses, explaining the overall hybrid improvement.

When we stratify by training frequency, the pattern becomes clearer. We categorize each test morpheme by its frequency in the training data: infrequent (1 to 5 occurrences), common (6 to 20), and frequent (over 20). Table~\ref{tab:error_frequency} shows accuracy by frequency bin.

\begin{table}[h]
\centering
\small
\begin{tabular}{lccc}
\toprule
\textbf{Frequency} & \textbf{Count} & \textbf{BiLSTM} & \textbf{Hybrid} \\
\midrule
Infrequent (1 to 5) & 69  & 0.029 & 0.426 \\
Common (6 to 20)    & 58  & 0.448 & 0.724 \\
Frequent (over 20)  & 86  & 0.860 & 0.859 \\
\bottomrule
\end{tabular}
\caption{Accuracy by training frequency for lexical morphemes. Hybrid improvements concentrate in infrequent morphemes (+0.397), with no test morphemes completely unseen in training.}
\label{tab:error_frequency}
\end{table}

Infrequent morphemes show dramatic improvement: BiLSTM achieves only 0.029 accuracy while the hybrid reaches 0.426 (+0.397). Common morphemes improve from 0.448 to 0.724 (+0.276). Frequent morphemes remain stable around 0.86 for both approaches. Notably, no test morphemes are completely unseen in this split. We did not enforce this property; with a document-level split and a small evaluation set, all test morphemes appear at least once in training. This demonstrates that hybrid gains stem from contextual inference on low-frequency items rather than handling zero-shot vocabulary, and that the baseline already captures frequent grammatical markers effectively when sufficient training examples exist.

\subsection{Key Findings Summary}
Our experiments reveal several important patterns for LLM-assisted morphological glossing. Retrieval-augmented prompting proves essential across all models, with similarity-based example selection dramatically outperforming random selection. Perhaps most surprisingly, providing morpheme dictionaries generally hurts performance: partial dictionaries universally degrade accuracy, while even complete dictionaries fail to help most models (only one of four shows gains). Performance scales approximately logarithmically with the number of in-context examples, typically peaking around ten to fifteen examples before plateauing or declining.

The hybrid architecture combining BiLSTM predictions with LLM correction improves performance across all tested models. These gains are particularly pronounced in low-shot scenarios where few examples are available, suggesting that the structured model provides valuable guidance when in-context learning is limited. Even without any examples, LLMs can successfully identify and correct many errors in structured predictions, indicating inherent morphological reasoning capabilities.

\section{Discussion}

\subsection{The Case for Hybrid Architectures}
Our results demonstrate that combining structured prediction with LLM reasoning yields consistent improvements across all tested models. The BiLSTM-CRF captures frequent morphological patterns from limited training data but struggles with rare morphemes and novel combinations. RAG LLM generation leverages broader linguistic knowledge and few-shot generalization but varies widely in accuracy depending on model choice. 

The hybrid pipeline combines these complementary strengths through a division of labor revealed by error analysis. Structured models are more accurate on grammatical morphemes, which likely reflects their high frequency and regularity in the training data, while LLMs excel at contextual inference for infrequent lexical items, leveraging retrieved examples to handle vocabulary gaps. Hybrid improvements concentrate in low-frequency morphemes where BiLSTM lacks sufficient training signal, while both approaches perform similarly on frequent items. This asymmetry explains why the two-stage architecture succeeds: each component addresses the other's primary weakness. BiLSTM predictions provide structural guidance (particularly valuable in low-shot scenarios) while LLMs refine these predictions using patterns from retrieved examples. Our prompt design frames the BiLSTM output as a fallible hypothesis rather than an authoritative baseline, encouraging critical evaluation while providing useful structural constraints.

While we do not claim that BiLSTM-CRF represents the optimal base model for this task, our results suggest a broader principle: combining any trainable structured predictor with retrieval-augmented LLM post-correction can yield gains over either approach alone. This synergy proves universally beneficial across all performance tiers, with particularly strong gains when in-context learning is limited by few examples or budget constraints. 

\subsection{The Role of Morpheme Dictionaries}
Perhaps our most surprising finding is that providing morpheme dictionaries generally hurts performance compared to providing no dictionary at all. Partial glossaries universally degrade accuracy, while even complete dictionaries fail to help most models. Only one model (\texttt{gemma-3-27b-it}) shows substantial gains from the complete dictionary, while three others perform worse with it than without.

These patterns suggest issues with how models integrate dictionary information, though we lack direct analysis of whether models actually consult dictionary entries. The degradation could stem from information overload, inappropriate dictionary consultation, or suboptimal prompt structure. Our approach provides dictionaries as simple unstructured key-value lists; alternative strategies merit investigation, such as organizing entries by morphological class or presenting dictionaries in structured formats.

Importantly, prompt structure can dramatically affect model behavior, and we have not systematically explored alternative formulations. Our findings should therefore be interpreted as demonstrating the ineffectiveness of our specific prompt design for dictionary integration, rather than fundamental limitations of dictionary use in morphological glossing. Future work should conduct systematic prompt engineering studies to identify more effective strategies for incorporating lexical resources.

\subsection{Practical Implications for Fieldwork}
Our findings suggest several considerations for practitioners working with LLM-assisted IGT annotation, though these patterns may vary across different languages, models, and documentation contexts. In our experiments, similarity-based retrieval consistently outperformed random example selection across all tested models, suggesting that investment in retrieval infrastructure may be worthwhile. The hybrid approach combining structured models with LLM post-correction showed gains across all configurations we tested, indicating potential value in this two-stage strategy.

For morpheme dictionaries, our results suggest caution: simple key-value presentations tended to hurt performance for most models in this setting, though alternative presentation strategies remain unexplored. Regarding few-shot example count, we observed approximately logarithmic scaling with diminishing returns beyond 10--15 examples, though optimal operating points likely depend on task complexity, model capabilities, and cost constraints. These patterns emerged from our specific experimental setup with Tuvan and four general-purpose LLMs; practitioners should validate these findings against their own languages and workflows, as model capabilities and architectural designs continue to evolve rapidly.

\subsection{Ethical Considerations}
Our experiments use data collected with informed consent under agreements restricting raw material sharing. We report only aggregate statistics and anonymized examples to protect speaker privacy. Speaker communities are not monolithic, and Tuvan speakers are distributed across multiple countries with different sociopolitical contexts. We do not claim community-wide consent; we follow the data agreements and consult the relevant fieldwork partners. Given the cross-border context, we avoid releasing raw data and refrain from identifying individuals or locations. Commercial APIs raise concerns about training data provenance \citep{bender_dangers_2021, sainz_nlp_2023}, and even at achieved accuracy levels, uncritical adoption risks introducing errors into the linguistic record. Decisions about automation should be made collaboratively with stakeholders, balancing efficiency gains against concerns about data control and quality.

\subsection{Future Directions}
Key directions include cross-linguistic evaluation on diverse morphological systems (polysynthetic, templatic, tonal) to test whether our design principles generalize beyond Turkic agglutination. Joint segmentation and glossing would address the limitation that our pipeline assumes gold boundaries. Parameter-efficient fine-tuning could improve performance with minimal language-specific data. Interactive interfaces with confidence estimation would help annotators prioritize review of uncertain predictions, and paradigm-level evaluation would better assess morphological generalization beyond token-level accuracy.

\section{Conclusion}
We present a hybrid automatic glossing pipeline combining BiLSTM-CRF structured prediction with LLM post-correction, evaluated on Tuvan fieldwork data across four LLMs. The hybrid approach consistently improves performance across all tested models, with particularly strong gains in low-shot scenarios where structural guidance from the BiLSTM proves most valuable. Error analysis reveals that improvements concentrate in lexical morphemes, especially rare vocabulary items, while BiLSTM already captures grammatical paradigms effectively. This demonstrates complementary strengths rather than simple performance stacking.

Our ablation studies reveal key design principles: retrieval-augmented prompting provides substantial gains; morpheme dictionaries generally hurt performance for most models; performance scales logarithmically with examples, plateauing around ten to fifteen; and hybrid correction benefits all models universally. These findings challenge conventional assumptions about prompt engineering, showing that more information is not always better.

While the utility of these accuracy levels for practical annotation workflows remains an open question dependent on community priorities and annotation contexts, the substantial error rates necessitate careful human oversight. Model outputs should be treated as hypotheses requiring expert validation rather than authoritative annotations.

\section{Limitations}
Our evaluation focuses on a single language (Tuvan) from one language family (Turkic), leaving generalization to other morphological systems untested. The patterns we observe may not hold for polysynthetic, templatic, or non-concatenative morphology. Additionally, our test set is small, reflecting realistic annotation costs but introducing sampling variance that limits statistical precision.

Our evaluation metrics capture only token-level accuracy rather than higher-order properties like paradigm consistency, morphophonological regularity, or alignment with community language priorities. The pipeline also assumes gold morpheme boundaries and does not address the segmentation problem, which itself requires substantial linguistic expertise in fieldwork contexts. 

The prompt engineering component of our study lacks systematic exploration. We tested only one prompt structure for each experiment, leaving unexplored how alternative formulations might affect dictionary effectiveness, example integration, or instruction following. Prompt design choices (information ordering, instruction phrasing, formatting conventions, and the balance of different knowledge sources) can dramatically impact model behavior, yet we lack the controlled comparisons needed to isolate their effects. Our glossary ablation findings in particular should be interpreted as showing that our specific prompt design failed to effectively leverage dictionary information, rather than demonstrating that dictionaries cannot help morphological glossing. We also lack fine-grained analysis of whether models actually consult dictionary entries or become distracted by additional prompt content.

Commercial API access limits transparency about training data and potential contamination from existing Tuvan linguistic resources. We use general-purpose LLMs rather than specialized models like GlossLM \citep{ginn_glosslm_2024}, which are explicitly trained on IGT data and may achieve higher absolute accuracy. However, our focus on establishing design principles (retrieval strategies, glossary configurations, hybrid architectures) likely generalizes across model types. Moreover, general-purpose LLMs offer practical advantages for fieldwork: no local infrastructure requirements, operation in few-shot regimes with minimal language-specific data, and accessibility to linguists without machine learning expertise. The tradeoff between specialized performance and practical accessibility merits further investigation.

\bibliography{custom}

\appendix

\section{Experimental Prompt Templates}
\label{sec:appendix}

This appendix provides the complete prompt templates used across all five experiments. All prompts follow a consistent structure with a system message defining the linguistic expert role, followed by task-specific instructions and formatting requirements.

\subsection{System Message (All Experiments)}
All experiments use the following system message:

\begin{quote}
\textit{You are a linguistic expert specializing in morpheme-by-morpheme glossing for an unknown language.}
\end{quote}

\subsection{Experiments 1--3: RAG LLM generation}
\label{sec:prompt-pure-llm}

Experiments 1 (Retrieval vs.\ Random), 2 (Glossary Ablation), and 3 (N-Shot Scaling) use RAG LLM generation with the following template:

\begin{quote}
\textbf{User message:}

Here are some examples of sentences with morpheme boundaries (marked by hyphens) and their glosses:

\textit{[For each example i in 1..k:]}

Example \textit{i}:

Segmented: \textit{[morpheme-segmented source text]}

Gloss: \textit{[corresponding gloss sequence]}

\textit{[If glossary provided:]}

You are also given a morpheme dictionary mapping morphemes to their English glosses. For morphemes in the dictionary, use the provided gloss. For others, infer from context. Some morphemes may have multiple translations; choose the most appropriate for this context.

Morpheme dictionary: \textit{[morpheme1: gloss1, morpheme2: gloss2, ...]}

\textit{[End glossary section]}

Please gloss this sentence:

Segmented: \textit{[test sentence with morpheme boundaries]}

Output the gloss with the same structure (spaces between words, hyphens between morphemes). Enclose your gloss in \#\#\#. Example: \#\#\#word1-MORPH1 word2-MORPH2\#\#\#
\end{quote}

\textbf{Experiment-specific variations:}
\begin{itemize}
\item \textbf{Experiment 1:} Uses 3 examples, no glossary. Compares TF-IDF retrieval (RAG) against random sampling.
\item \textbf{Experiment 2:} Uses RAG, no glossary. Varies $k \in \{1, 3, 5, 10, 15, 20\}$ examples.
\item \textbf{Experiment 3:} Uses 3 RAG examples. Tests four glossary sizes: none (0 pairs), top-100 (100 pairs), grammatical (240 pairs), entire (1,498 pairs).
\end{itemize}

\subsection{Experiment 4: Hybrid Pipeline (BiLSTM + LLM)}
\label{sec:prompt-hybrid}

Experiment 4 uses BiLSTM-CRF predictions as initial hypotheses for LLM correction, with the following template:

\begin{quote}
\textbf{User message:}

You will be given:
\begin{enumerate}
\item A rough initial glossing attempt from a statistical model (may contain errors)
\item Some example sentences with their correct glosses
\item A morpheme dictionary
\end{enumerate}

Your task is to produce the correct gloss, using all available information.

Here are some example sentences with correct glosses:

\textit{[For each example i in 1..k (or 0 for zero-shot):]}

Example \textit{i}:

Segmented: \textit{[morpheme-segmented source text]}

Gloss: \textit{[corresponding gloss sequence]}

\textit{[If glossary provided:]}

You also have access to a morpheme dictionary: \textit{[morpheme1: gloss1, morpheme2: gloss2, ...]}

\textit{[End glossary section]}

Now, please gloss this sentence:

Segmented: \textit{[test sentence with morpheme boundaries]}

Initial attempt (from statistical model): \textit{[BiLSTM-CRF prediction]}

This initial attempt may contain errors. Use the examples, dictionary, and linguistic patterns to produce the correct gloss. Maintain the same structure (spaces between words, hyphens between morphemes).

IMPORTANT: Output ONLY the gloss wrapped in \#\#\#. Do not explain your reasoning. Example format: \#\#\#word1-MORPH1 word2-MORPH2\#\#\#
\end{quote}

\textbf{Design rationale:} The hybrid prompt presents the BiLSTM prediction as a ``rough initial attempt'' rather than an authoritative baseline, encouraging the LLM to critically evaluate and correct errors. Clean gold examples (not error-correction pairs) provide paradigmatic context, while the statistical prediction narrows the search space by proposing plausible morpheme boundaries and candidate glosses.

\subsection{Output Extraction}
All experiments extract model outputs by locating text between \texttt{\#\#\#} delimiters. If delimiters are absent (indicating non-compliance), the entire output string is used as the predicted gloss. This extraction method proved robust across all four LLM providers.






















\section{Detailed Results Tables}
\label{sec:appendix-tables}

This section provides detailed numerical results for experiments presented as figures in the main text.

\begin{table}[h]
\centering
\small
\begin{tabular}{lcccc}
\toprule
\textbf{Glossary} & \textbf{deepseek} & \textbf{qwen3} & \textbf{gpt-4o} & \textbf{gemma-3} \\
\midrule
None          & \textbf{0.510} & \textbf{0.436} & \textbf{0.420} & 0.325 \\
Top-100     & 0.469 & 0.332 & 0.353 & 0.430 \\
Grammatical & 0.426 & 0.303 & 0.364 & 0.335 \\
Entire     & 0.490 & 0.422 & 0.369 & \textbf{0.455} \\
\bottomrule
\end{tabular}
\caption{Glossary ablation study across four LLMs (3-shot RAG). Partial glossaries consistently degrade performance, while complete dictionaries fail to help most models. Model names abbreviated: deepseek = \texttt{deepseek-v3.2-exp}, qwen3 = \texttt{qwen3-max}, gpt-4o = \texttt{gpt-4o-mini}, gemma-3 = \texttt{gemma-3-27b-it}.}
\label{tab:glossary_all_models}
\end{table}

\begin{table}[h]
\centering
\small
\begin{tabular}{ccccc}
\toprule
\textbf{N} & \textbf{deepseek} & \textbf{qwen3} & \textbf{gpt-4o} & \textbf{gemma-3} \\
\midrule
1  & 0.350 & 0.196 & 0.159 & 0.118 \\
3  & 0.505 & 0.439 & 0.383 & 0.336 \\
5  & 0.588 & 0.519 & 0.423 & 0.471 \\
10 & \textbf{0.658} & 0.529 & 0.430 & \textbf{0.600} \\
15 & 0.646 & \textbf{0.611} & \textbf{0.486} & 0.540 \\
20 & 0.626 & 0.552 & 0.475 & 0.534 \\
\bottomrule
\end{tabular}
\caption{N-shot scaling for RAG LLM generation (RAG, no glossary). Performance improves logarithmically, with diminishing returns beyond n=10--15. Model names abbreviated as in Table~\ref{tab:glossary_all_models}.}
\label{tab:nshot_scaling}
\end{table}

\end{document}